\documentclass[10pt,twocolumn,letterpaper]{article}

\usepackage{cvpr}
\usepackage{times}
\usepackage{epsfig}
\usepackage{graphicx}
\usepackage{amsmath}
\usepackage{amssymb}
\usepackage{comment}
\usepackage[normalem]{ulem}
\usepackage{algorithm}
\usepackage{algorithmic}
\usepackage{subfigure}
\usepackage{graphics}
\usepackage{stfloats}

\usepackage[pagebackref=true,breaklinks=true,letterpaper=true,colorlinks,bookmarks=false]{hyperref}

\cvprfinalcopy 

\def\cvprPaperID{553} 

\ifcvprfinal\fi
\begin{document}

\title{Low Rank Representation on Grassmann Manifolds: An Extrinsic Perspective}

\author{Boyue Wang$^{1}$, Yongli Hu$^{1}$, Junbin Gao$^{2}$, Yanfeng Sun$^{1}$ and Baocai Yin$^{1}$\\[1mm]
$^{1}$Beijing Key Laboratory of Multimedia and Intelligent Software Technology\\
College of Metropolitan Transportation, Beijing University of Technology\\
Beijing, 100124, China\\
boyue.wang@gmail.com, \{huyongli,yfsun,ybc\}@bjut.edu.cn
\and
$^2$School of Computing and Mathematics\\
Charles Sturt University\\
Bathurst, NSW 2795, Australia\\
jbgao@csu.edu.au}

\maketitle
 
\begin{abstract}
Many computer vision algorithms employ subspace models to represent data.
The Low-rank representation (LRR) has been successfully applied in subspace clustering for which data are clustered according to their subspace structures.  The possibility of extending LRR on Grassmann manifold is explored in this paper. Rather than directly embedding Grassmann manifold into a symmetric matrix space, an extrinsic view is taken by building the self-representation of LRR over the tangent space of each Grassmannian point. A new algorithm for solving the proposed Grassmannian LRR model is designed and implemented. Several clustering experiments are conducted on handwritten digits dataset, dynamic texture video clips and YouTube celebrity face video data. The experimental results show our method outperforms a number of existing methods.
\end{abstract}

\section{Introduction}

In modern computer vision research, one always models a collection of data samples as a subspace. For example, the set of handwritten digital images of a single digit from a particular person can be modeled as a subspace, which can be defined by several top principal components revealed by PCA (principal component analysis) \cite{WangHuGaoSunYin2014}. Alternatively, a video clip can be viewed as a collection of data samples (frames), thus it is natural to regard the video clip for a particular scene/action as a whole to be modeled as the subspace that spans the observed frames \cite{HarandiSalzmannJayasumanaHartleyLi2014,TuragaVeeraraghavanSrivastavaChellappa2011}. The \textit{data samples} produced from video data or image sets in this way are in fact the subspaces, rather than the data points in Euclidean spaces.

Our main motivation here is to develop new methods for analysing video data and image sets through subspace representations, by borrowing the ideas used in analysing data samples in linear spaces. It should be mentioned that the clustering problem to be researched in this paper is different from the problem in the conventional subspace clustering where the fundamental purpose is to cluster data samples in the whole Euclidean spaces into clusters according to their subspace structures. That is, the objects to be clustered are samples in the main linear space while in this paper the objects themselves are subspaces (of the same dimension), i.e., the points on the abstract Grassmann manifold \cite{AbsilMahonySepulchre2004,EdelmanAriasSmith1998}. The relevant research can be seen in e.g. \cite{CetingulVidal2009}.

The subspace clustering has attracted great interest in computer vision, pattern recognition and signal processing \cite{ElhamifarVidal2013,WrightMaMairalSapiroHuang2010,XuWunsch-II2005,ZhangGhanemLiuAhuja2012}.  Vidal \cite{Vidal2011} classified subspace clustering algorithms into four approaches: algebraic \cite{HongWrightHuangMa2006,Kanatani2001,MaYangDerksenFossum2008}, statistical \cite{GruberWeiss2004,TippingBishop1999a}, iterative \cite{HoYangLimLeeKriegman2003,Tseng2000} and spectral clustering based \cite{ElhamifarVidal2013,LangLiuYuYan2012,LiuLinSunYuMa2013}. The procedure of spectral clustering based methods consists of two steps: (1) learning a similarity matrix for the given data sample set; and (2) performing spectral clustering to categorize data samples such as K-means or Normalized Cuts (NCut) \cite{ShiMalik2000}.

Two representatives of spectral clustering based methods are Sparse Subspace Clustering (SSC) \cite{ElhamifarVidal2013} and Low Rank Representation (LRR) \cite{LiuLinSunYuMa2013}.
Both SSC and LRR rely on the self expressive property in linear space \cite{ElhamifarVidal2013}: \textit{each data point in a union of subspace can be efficiently reconstructed by a combination of other points in the data.}
SSC further induces sparsity by utilizing the $l_1$ Subspace Detection Property \cite{Donoho2004} in an independent manner, while LRR model considers the intrinsic relation among the data objects in a holistic way via the low rank requirement. It has been proved that, when the high-dimensional data set is actually composed of a union of several low dimension subspaces, LRR model can reveal this structure through subspace clustering \cite{LiuLinYu2010}.

However the principle of self expression is only valid and applicable in classic linear spaces. As mentioned above, the data samples on Grassmann manifold are abstract.  It is well known \cite{Chikuse2002a} that Grassmann manifold is isometrically equivalent to the subspace of symmetric idempotent
matrices. To get around the difficulty of SSC/LRR self expression in abstract Grassmann setting, the authors of \cite{WangHuGaoSunYin2014} embed the Grassmann manifold into the symmetric matrix manifold where the self expression can be naturally defined, thus an LRR model on Grassmann manifold was formulated.

In this paper, we review the geometric properties of Grassmann manifold and take an extrinsic view to develop a new LRR or SSC model by exploring tangent space structure of Grassmann manifold.  The new formulation relies on the so-called Log Mapping on Grassmann manifold through which the self expression of Grassmann points can be implemented in the tangent space at each particular Grassmann point. By this way, we extend the classic LRR model onto Grassmann manifold, namely GLRR, so LRR can be used for representing the non-linear high dimensional data and implement clustering on the manifold space.

The primary contribution of this paper is to
\begin{enumerate}
\item formulate the LRR model on Grassmann Manifold by exploring the linear relation among all the mapped points in the tangent space at each Grassmann point. The mapping is implemented by the Log mapping on the Grassmann manifold; and
\item propose a practical algorithm for the solution to the problem of the extended Grassmann LRR model.
\end{enumerate}

The rest of the paper is organized as follows. In Section \ref{Sec:2}, we summarize the related works and the preliminaries for Grassmann manifold. Section \ref{Sec:3} describes the low rank representation on Grassmann Manifold. In Section \ref{Sec:4}, an algorithm for solving the LRR model on Grassmann manifold is proposed. In Section \ref{Sec:5}, the performance of the proposed method is evaluated on clustering problems of three public databases. Finally, conclusions and suggestions for future work are provided.

\section{Preliminaries and Related Works}\label{Sec:2}

In this section, first, we briefly review the existing sparse subspace clustering methods, SSC and LRR. Then we describe the fundamentals of Grassmann manifold which are relevant to our proposed model and algorithm.

\subsection{SSC \cite{ElhamifarVidal2013} and LRR \cite{LiuLinSunYuMa2013}}\label{Subsec:2.1}
 
Given a set of data drawn from an unknown union of subspaces $\mathbf{X} = [\mathbf x_1, \mathbf x_2, ..., \mathbf x_N]\in\mathbb{R}^{D\times N}$ where $D$ is the data dimension, the objective of subspace clustering is to assign each data sample to its underlying subspace. The basic assumption is that the data in $\mathbf{X}$ are drawn from union of $K$ subspaces $\{\mathcal{S}_k\}^K_{k=1}$ of dimensions $\{d_k\}^K_{k=1}$.

Under the data self expressive principle, it is assumed that each point in data can be written as a linear combination of other points i.e., $\mathbf{X} = \mathbf{X}\mathbf{Z}$, where $\mathbf{Z}\in\mathbb{R}^{N\times N}$ is a matrix of similarity coefficients. In the case of corrupted data, we may relax the self expressive model to $\mathbf{X} = \mathbf{X}\mathbf{Z}+\mathbf E$, where $\mathbf E$ is a fitting error. With this model, the main purpose is to learn the expressive coefficients $\mathbf Z$, under some appropriate criteria, which can be used in a spectral clustering algorithm.

The optimization criterion for learning $\mathbf Z$ is
\begin{align}
\min_{\mathbf Z, \mathbf E}\|\mathbf E\|^2_p + \lambda \|\mathbf Z\|_q,  \text{ s.t. } \mathbf{X} = \mathbf{X}\mathbf{Z}+\mathbf E,\label{BasicSparseModel}
\end{align}
where $p$ and $q$ are norm place-holders and $\lambda>0$ is a penalty parameter to balance the coefficient term and the reconstruction error.

The norm $\|\cdot\|_p$ is used to measure the data expressive errors while the norm $\|\cdot\|_q$ is chosen as a sparsity inducing measure. The popular choices for $\|\cdot\|_p$ is Frobenius norm $\|\cdot\|_F$ (or $\|\cdot\|_2$). SSC and LRR differentiate at the choice of $\|\cdot\|_q$.  SSC is in favour of the sparsest representation for the data sample using $\ell_1$ norm \cite{Vidal2011} while LRR takes a holistic view in favour of a coefficient matrix in the lowest rank, measured by the nuclear norm $\|\cdot\|_*$.

To avoid the case in which a data sample is represented by itself, an extra constraint $\text{diag}(\mathbf Z)=0$ is included in SSC model. LRR uses the so-called $\ell_1/\ell_2$ norm to deal with random gross errors in data. In summary, we re-write SSC model as follows,
\begin{align}
\min_{\mathbf Z, \mathbf E}\|\mathbf E\|^2_F + \lambda \|\mathbf Z\|_1,  \text{ s.t. } \mathbf{X} = \mathbf{X}\mathbf{Z}+\mathbf E, \text{diag}(\mathbf Z) = 0,\label{SSCModel}
\end{align}
and LRR model
\begin{align}
\min_{\mathbf Z, \mathbf E}\|\mathbf E\|^2_{1/2} + \lambda \|\mathbf Z\|_*,  \text{ s.t. } \mathbf{X} = \mathbf{X}\mathbf{Z}+\mathbf E.\label{LRRModel}
\end{align}
 
\subsection{Grassmann Manifold}
This paper is concerned with points particularly on a known manifold. In most cases in computer vision applications, manifolds can be considered as low dimensional smooth ``surfaces'' embedded in a higher dimensional Euclidean space. A manifold is locally similar to Euclidean space around each point of the manifold.

In recent years, Grassmann manifold \cite{BounmalAbsil2014} has attracted great interest in research community, such as subspace tracking \cite{SrivastavaKlassen2004}, clustering \cite{CetingulVidal2009}, discriminant analysis \cite{HammLee2008}, and sparse coding \cite{HarandiSandersonShenLovell2013,HarandiSandersonShiraziLovell2011}. Mathematically Grassmann manifold $\mathcal{G}(p,d)$ is defined as the set of $p$-dimensional subspaces in $\mathbb{R}^d$. Its Riemannian geometry has been recently well investigated in literature \cite{AbsilMahonySepulchre2008,AbsilMahonySepulchre2004,EdelmanAriasSmith1998}.

As a subspace in $\mathcal{G}(p,d)$ is an abstract concept, a concrete representation must be chosen for numerical learning purposes. There are several classic concrete representations for the abstract Grassmann manifold in literature. For our purpose, we briefly list some of  them. Denote by $\mathbb{R}^{d\times p}_*$ the space of all $d\times p$ matrix of full column rank; $GL(p)$ the general group of nonsingular matrices of order $p$; $\mathcal{O}(p)$ the group of all the $p\times p$ orthogonal matrices.

\textit{\uppercase\expandafter{\romannumeral1}. Representation by Full Columns Rank Matrices \cite{AbsilMahonySepulchre2008}:}
\[
\mathcal{G}(p,d) \cong \mathbb{R}^{d\times p}_* / GL(p)
\]

\textit{\uppercase\expandafter{\romannumeral2}. The Orthogonal Representation \cite{EdelmanAriasSmith1998}:}
\[
\mathcal{G}(p,d) \cong \mathcal{O}(d) / \mathcal{O}(p) \times \mathcal{O}(d-p)
\]
 
\textit{\uppercase\expandafter{\romannumeral3}. Symmetric Idempotent Matrix Representation \cite{Chikuse2002a}:}
\[
\mathcal{G}(p,d) \cong \{\mathbf P\in\mathbb{R}^{d\times d}:  \mathbf P^T = \mathbf P, \mathbf P^2=\mathbf P, \text{rank}(\mathbf P) = p\}.
\]
Many researchers adopt this representation for learning tasks on Grassmann manifold, for example, Grassmann Discriminant Analysis \cite{HammLee2008}, Grassmann Kernel Methods \cite{HarandiSalzmannJayasumanaHartleyLi2014}, Grassmann Low Rank Representation \cite{WangHuGaoSunYin2014}, etc.

\textit{\uppercase\expandafter{\romannumeral4}. The Stiefel Manifold Representation \cite{EdelmanAriasSmith1998}:}
Denote by $\mathcal{ST}(p,d) = \{\mathbf X\in\mathbb{R}^{d\times p}: \mathbf X^T\mathbf X = I_p\}$ the set of all the $p$ dimension bases, called Stiefel manifold. We identify a point $\mathcal{S}$ on Grassmann manifold $\mathcal{G}(p,d)$ as an equivalent class under orthogonal transform of Stiefel manifold. Generally we have
\[
\mathcal{G}(p,d) \cong \mathcal{ST}(p,d) / \mathcal{O}(p).
\]

Two bases $\mathbf X_1$ and $\mathbf X_2$ are equivalent if there exist an orthogonal matrix $\mathbf Q\in \mathcal{O}(p)$ such that $\mathbf X_1 = \mathbf X_2\mathbf Q$.

In this paper, we will work on the Stiefel Manifold Representation for Grassmann manifold.  Each point on the Grassmann manifold is an equivalent class defined by
\begin{align}
[\mathbf X]  = \{\mathbf X\mathbf Q | \mathbf X^T\mathbf X = I_p, \mathbf Q \in\mathcal{O}(p)\} \label{EquivalentClass}
\end{align}
where $\mathbf X$, a representative of the point $[X]$, is a point on the Stiefel manifold, i.e., a matrix of $\mathbb{R}^{d\times p}$ with orthogonal columns.

\section{LRR on Grassmann Manifold}\label{Sec:3}
\subsection{The Idea and the Model}
The famous \textit{Science} paper by Roweis and Saul \cite{RoweisSaul2000} proposes a manifold learning algorithm for low dimension space embedding by approximating manifold structure of data. The method is named locally linear embedding (LLE). Similarly a latent variable version, Local Tangent Space Alignment (LTSA), can be seen in \cite{ZhangZha2004}.

Both LLE and LTSA exploit the ``manifold structures'' implied by data. In this setting, we have no knowledge about the manifold where the data reside. However in many computer vision tasks, we know the manifold where the data come from. The idea of using manifold information to assist in learning task can be seen in earlier research work \cite{LeeAbbottAraman2009} and the recent work \cite{HarandiSalzmannHartley2014}.

To construct the LRR model on manifolds, we first recall LLE algorithm. LLE relies on learning the local linear combination $\mathbf X_i \approx \sum_{j} w_{ij}\mathbf X_j$ (here we use all the other data as the neighbours of $\mathbf X_i$ for convenience) under the condition $\sum_{j} w_{ij} = 1$ (assume $w_{ii}=0$). Hence
\begin{align}
\sum_{j} w_{ij}\mathbf X_j - \mathbf X_i = \sum_{j}w_{ij}(\mathbf X_j - \mathbf X_i) \approx 0. \label{TangentApproximation}
\end{align}

We note that in Euclidean case $\mathbf X_{j}-\mathbf X_i$ can be regarded as an approximated tangent vector at point $\mathbf X_i$. Eqn \eqref{TangentApproximation} means that the combination of all these tangent vectors should be close to 0 vector. It is well known that on general manifold the Euclidean tangent $\mathbf X_{j}-\mathbf X_i$ at $\mathbf X_i$  can be replaced by the result of the Log mapping $\text{Log}_{\mathbf X_i}(\mathbf X_j)$ which maps $\mathbf X_j$ on a manifold to a tangent vector at $\mathbf X_i$. Thus the data self expressive principle used in \eqref{BasicSparseModel} can be realized on the tangent space of each data as
\[
\sum_{j}w_{ij}\text{Log}_{\mathbf X_i}(\mathbf X_j) \approx 0.
\]

This idea of moving linear relation over to tangent space has been used in \cite{CetingulWrightThompsonVidal2014,GohVidal2008}. The above request was also obtained in \cite{XieHoVemuri2013} by the approximately defined linear combination on manifold for the purpose of dictionary learning over Riemannian manifolds.

Finally we formulate the SSC/LRR on manifold as the following optimization problem:
\begin{align}
\begin{aligned}
&\min_{W}\sum^N_{i=1}\bigg\|\sum^N_{j=1}w_{ij}\text{Log}_{X_i}(X_j)\bigg\|^2_{X_i} + \lambda \|W\|_q  \\
&\text{s.t.} \sum^N_{j=1}w_{ij} = 1,  i=1,2,..., N,
\end{aligned}\label{25August2014-3}
\end{align}
where $\text{Log}_{\cdot}(\cdot)$ is defined as the Log mapping on the manifold and $\sum^N_{j=1}w_{ij} = 1$ are affine constraints to preserve the coordinate independence on manifold same as the term of the nonlinear sparse representation in \cite{XieHoVemuri2013}. When $\|\cdot\|_q$ is $\|\cdot\|_1$, we have SSC model on manifolds, and when $\|\cdot\|_q$ takes $\|\cdot\|_*$, LRR on manifolds is formed. In the sequel, we mainly focus on LRR on Grassmann manifold.

\subsection{Properties of Grassmann Manifold Related to LRR Learning}
For the special case of Grassmann manifold, formulation \eqref{25August2014-3} is actually abstract. We need to specify the meaning of the norm $\|\cdot\|_{\mathbf X}$ and $\text{Log}_{\mathbf X}(\mathbf Y)$.
\subsubsection{Tangent Space and Its Metric}
Consider a point $\mathcal{S} = \text{span}(\mathbf X)$ on Grassmann manifold $\mathcal{G}(p,d)$ where $\mathbf X$ is a representative of the equivalent class defined in \eqref{EquivalentClass}. We denote the tangent space at $\mathcal{S}$ by $T_{\mathcal{S}}\mathcal{G}(p,d)$.  A tangent vector $\mathcal{H}\in T_{\mathcal{S}}\mathcal{G}(p,d)$ is represented by a matrix $\mathbf H\in\mathbb{R}^{d\times p}$ verifying, see \cite{BounmalAbsil2014},
\[
\frac{d}{dt}\text{span}(\mathbf X+t\mathbf H)|_{t=0} = \mathcal{H}.
\]

Under the condition that $\mathbf X^T\mathbf H=0$, it can be proved that $\mathbf H$ is the unique matrix as the representation of tangent vector $\mathcal{H}$, known as its horizontal lift at the representative $\mathbf X$. Hence the abstract tangent space  $T_{\mathcal{S}}\mathcal{G}(p,d)$ at $\mathcal{S} = \text{span}(\mathbf X)$ can be represented by the following concrete set
\[
T_{\mathbf X}\mathcal{G}(p,d) = \{\mathbf H\in\mathbb{R}^{d\times p} |  \mathbf X^T\mathbf H = 0\}.
\]

The above representative tangent space is embedded in matrix space $\mathbb{R}^{d\times p}$ and it inherits the canonical inner
\begin{align*}
\forall \mathbf H_1, \mathbf H_2\in T_{\mathbf X}\mathcal{G}(p,d), \ \langle \mathbf H_1, \mathbf H_2\rangle_{\mathbf X} = \text{trace}(\mathbf H^T_1\mathbf H_2).  
\end{align*}
Under this metric, the Grassmann manifold is Riemannian and the norm term used in \eqref{25August2014-3} becomes
\begin{align}
\| \mathbf H \|^2_{\mathbf X}= \langle \mathbf H, \mathbf H\rangle_{\mathbf X} = \text{trace}(\mathbf H^T\mathbf H)
\label{Norm}
\end{align}
which is irrelevant to the point $\mathbf X$.  Note that $\mathbf H$ is a tangent vector at $\mathbf X$ on Grassmann manifold.

\subsubsection{Log Mapping of Grassmann Manifold} \label{ExpLog}
Log map on general Riemannian manifold is the inverse of the Exp map on the manifold \cite{Lee2002}. For Grassmann manifold, there is no explicit expression for the Log mapping. However the Log operation  can be written out by the following algorithm \cite{EdelmanAriasSmith1998,RentmeestersAbsilDooren2010}:

Given two representative Stiefel matrices $\mathbf X, \mathbf Y$ of equivalent classes $[\mathbf X], [\mathbf Y]\in\mathcal{G}(p, d)$ as points on the Grassmann manifold (matrix representations in $d$-by-$p$ matrices), we seek to find $\mathbf H$ at $\mathbf X$ such that exponential map $\exp_{\mathbf X} (\mathbf H) = \mathbf Y$.

Instead of $\mathbf H$, we equivalently identify its thin-SVD $\mathbf H = \mathbf U\mathbf S\mathbf V^T$. Let us consider these equations, where $\mathbf Y$ is obtained with the exponential map $\exp_{\mathbf X} (\mathbf U\mathbf S\mathbf V^T)$,
\[
\mathbf Y = \mathbf X \mathbf V \cos(\mathbf S) \mathbf Q^T + \mathbf U \sin(\mathbf S) \mathbf Q^T,
\]
where $\mathbf Q$ is  any $d$-by-$d$ orthogonal matrix.

Consequently, we need to solve the equations of $\mathbf U, \mathbf S, \mathbf V$, and $\mathbf Q$, with the knowledge that $\mathbf X^T \mathbf H = 0$, since $\mathbf H$ is in the tangent space,
\begin{align*}
\mathbf V \cos(\mathbf S) \mathbf Q^T = \mathbf X^T\mathbf Y, \\
\mathbf U \sin(\mathbf S) \mathbf Q^T = \mathbf Y - \mathbf X \mathbf X^T\mathbf Y.
\end{align*}
Hence
\[
\mathbf U \sin(\mathbf S) \mathbf Q^T (\mathbf V\cos(\mathbf S)\mathbf Q^T)^{-1} = (\mathbf Y - \mathbf X \mathbf X^T\mathbf Y)(\mathbf X^T\mathbf Y)^{-1}
\]
which gives
\[
\mathbf U \tan(\mathbf S) \mathbf V^T  = (\mathbf Y - \mathbf X \mathbf X^T\mathbf Y)(\mathbf X^T\mathbf Y)^{-1}
\]
because $\mathbf V$ is actually an orthogonal matrix.  Hence the algorithm will be conducting a SVD decomposition of $\mathbf U\Sigma \mathbf V^T = (\mathbf Y - \mathbf X \mathbf X^T\mathbf Y)(\mathbf X^T\mathbf Y)^{-1}$, then define
\begin{align}
\text{Log}_{\mathbf X}(\mathbf Y) = \mathbf H = \mathbf U \arctan(\Sigma) \mathbf V^T.
\label{LogGrassmann}
\end{align}

\section{Solution to GLLR}\label{Sec:4}
With the concrete specifications \eqref{Norm} and \eqref{LogGrassmann} for Grassmann manifold, the problem for LRR on Grassmann manifold \eqref{25August2014-3} becomes numerically computable.
Denote by
\begin{align}
B^i_{jk} = \text{trace}(\text{Log}_{\mathbf X_i}(\mathbf X_j)^T \text{Log}_{\mathbf X_i}(\mathbf X_k)), \label{B}
\end{align}
which can be calculated according to the Log algorithm as defined in \eqref{LogGrassmann} for all the given data $X_i$'s. Then problem \eqref{25August2014-3} can be rewritten as
\begin{align}
\begin{aligned}
&\min_{W}\sum^N_{i=1} \mathbf w_i B_i \mathbf w^T_i + \lambda \|W\|_*  \\
&\text{s.t.} \sum^N_{j=1}w_{ij} = 1,  i=1,2,..., N,
\end{aligned}\label{25August2014-4}
\end{align}
where $B_i = (B^i_{jk})$ and $\mathbf w_i = (w_{i1}, ..., w_{iN})$ is the row vector of $W$. 
The augmented Lagrangian objective function of \eqref{25August2014-4} is given by
\begin{align}
\begin{aligned}
L = &\sum^N_{i=1}\left\{\frac12\mathbf w_i B_i\mathbf w^T_i + y_i(\sum^N_{j=1}w_{ij}-1)\right. \\
 &+  \left.\frac{\beta_k}2(\sum^N_{j=1}w_{ij}-1)^2\right\} + \lambda \|W\|_*
\end{aligned}
\label{25August2014-5}
\end{align}
where $y_i$ are Lagrangian multipliers  and we have used an adaptive constant $\beta_k$.

Denote the first summation term in \eqref{25August2014-5} by $F(W)$. 
At the current location $W^{(k)}$, we take a linearization of $F$,
\begin{align*}
F(W)\approx & F(W^{(k)}) + \langle \partial F(W^{(k)}), W-W^{(k)}\rangle \\
&+ \frac{\eta_B\beta_k}2\|W-W^{(k)}\|^2_F
\end{align*}
where $\partial F(W^{(k)})$ is a gradient matrix with $i$-th row given by
\begin{align}
\mathbf w^{(k)}_i B_i + y^{(k)}_i\mathbf 1 + \beta_k(\sum_{j=1}w^{(k)}_{ij}-1)\mathbf 1 \label{Eq:14October2014-4}
\end{align}
with $\mathbf 1\in \mathbb{R}^N$ the row vector of all ones. Let
$\eta_B = \max\{\|B_i\|^2\}+N+1$.
 
Taking this into \eqref{25August2014-5}, we define the following iteration
\begin{align}
 W^{(k+1)}   
= &\arg\min_W \lambda \|W\|_* \label{SolutionW}\\
+&\frac{\eta_B\beta_k}{2} \bigg\|W - \left(W^{(k)} - \frac{1}{\eta_B\beta_k} \partial F(W^{(k)})\right)\bigg\|^2_F. \notag
\end{align}
Problem \eqref{SolutionW} admits a closed form solution by using SVD thresholding operator \cite{CaiCandesShen2008}.

The update rules for $y_i$ is
\begin{align}
y^{(k+1)}_i = y^{(k)}_i + \beta_k (\sum^N_{j=1}w^{(k)}_{ij}-1) \label{Eq:14October2014-5}
\end{align}
and the updating rule for $\beta$
\begin{align}
\beta_{k+1} = \min\{\beta_{\text{max}}, \rho \beta_k\}. \label{UpdateBeta}
\end{align}
where
 \[
 \rho = \begin{cases} \rho_0 & \beta_k \|W^{k+1} - W^k\| \leq \varepsilon_1\\
 1 & \text{otherwise}
 \end{cases}
 \]

We summarize the above as Algorithm 1.

\begin{algorithm}\label{wholeAlg1a}
\renewcommand{\algorithmicrequire}{\textbf{Input:}}
\renewcommand\algorithmicensure {\textbf{Output:} }
\caption{ Low-Rank Representation on Grassmann Manifold.}
\begin{algorithmic}[1]
\REQUIRE The Grassmann sample set $\{X_i\}_{i=1}^N$,$X_i\in \mathcal{G}(p,d)$ and the balancing parameter $\lambda$.  \\
\ENSURE  The Low-Rank Representation $Z$ ~~\\
\STATE   Initialize: Set the parameters $\rho_0=1.9$, $\eta_B = \max\{\|B_i\|^2\}+N+1$, $\beta_{\max}=10^6\gg \beta_0 = 0.1$, $\varepsilon_1 = 1e-4$, $\varepsilon_2 = 1e-4$, $W^0 = 0$, $\mathbf y_0=0$. \\
\STATE   Prepare all $\mathbf B_i$ according to \eqref{B}
\WHILE   {not converged}
\STATE   Calculate $\partial F(W^{(k)})$ row by row according to \eqref{Eq:14October2014-4}
\STATE   Let $M_k = W^{(k)} -  \frac1{\eta_B\beta_k} \partial F(W^{(k)})$, then
\STATE   Update the $W^{k+1}$ according to \eqref{SolutionW}
\STATE   Update $y^{k+1}$ by \eqref{Eq:14October2014-5}
\STATE   Update $\beta_{k+1}$ by \eqref{UpdateBeta}
\STATE   Check the convergence condition: $\beta_k \|W^{k+1} - W^k\| \leq \varepsilon_1$ and $\|W^{k+1}\mathbf 1 - \mathbf 1\| \leq \varepsilon_2$
\ENDWHILE
\end{algorithmic}
\end{algorithm}

\section{Experiments}\label{Sec:5}
In this section, we evaluate the performance of the proposed GLRR on three widely used public databases: (\romannumeral1) MNIST handwritten digits image set \cite{Yann1998}; (\romannumeral2) DynTex++ dynamic texture videos \cite{GhanemAhuja2010}; and (\romannumeral3) YouTube Celebrity faces videos \cite{KimKumarPavlovicRowley2008,WangGuoDavisDai2012}.
\subsection{Data Preparation and Experiment Settings}
\begin{figure}
    \begin{center}
    \includegraphics[width=0.45\textwidth]{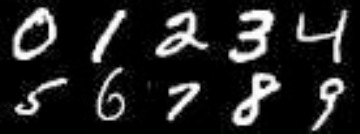}
    \end{center}
    \caption{The MNIST digit samples for experiments.}\label{MNISTfig}
\end{figure}

MNIST handwritten digits database \cite{Yann1998} consists of approximately 70,000 digit images written by 250 volunteers. All the digit images have been size-normalized and centered in a fixed size of $28\times 28$. This data set can be regarded as synthetic as the data are clean. Some samples of this database are shown in Figure \ref{MNISTfig}. We use this data set to test the clustering performance of the proposed method as noise-free cases.

\begin{figure}
    \begin{center}
    \includegraphics[width=0.45\textwidth,height=25mm]{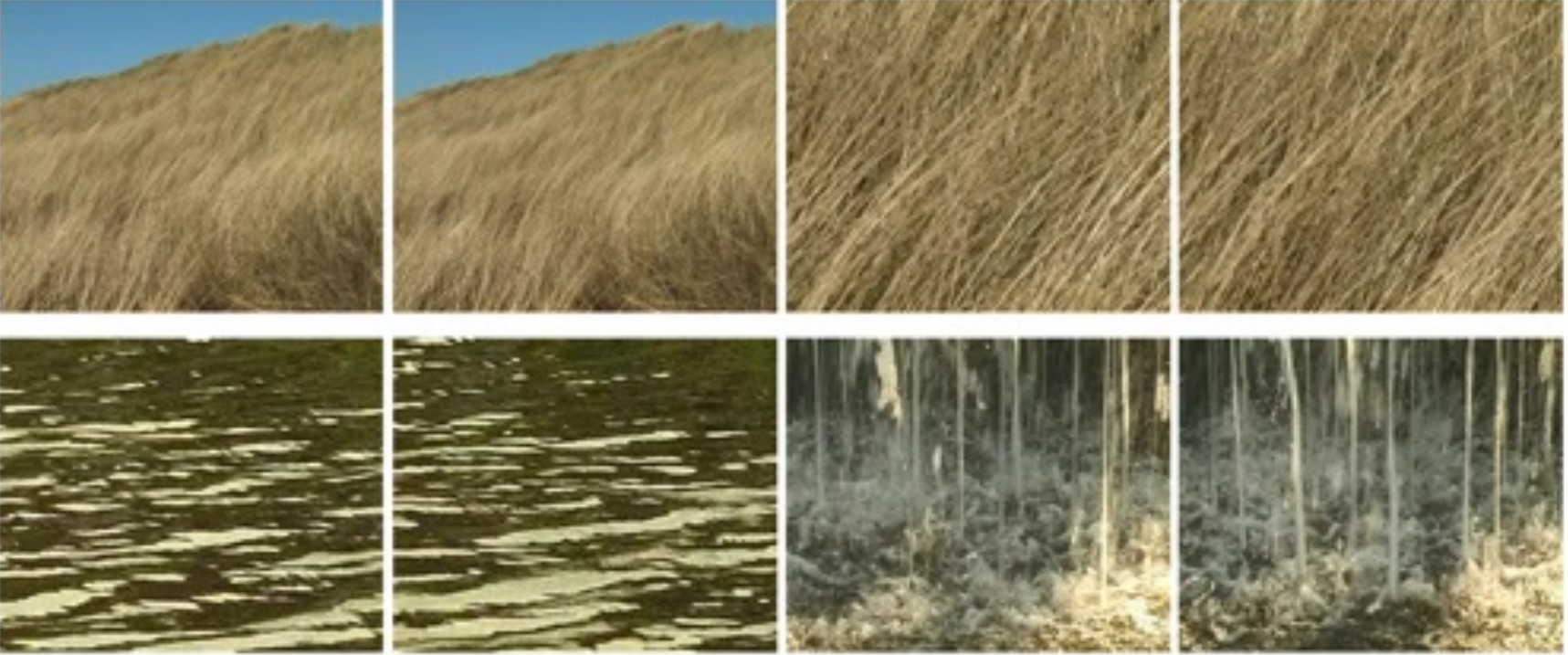}
    \end{center}
    \caption{DynTex++ samples. each two frames from same video sequence.}\label{DynTexfig}
\end{figure}

To test the clustering performance of the proposed method for multiple classes, we select DynTex++ database \cite{GhanemAhuja2010}. The dataset is derived from a total of 345 video sequences in different scenarios which contains river water, fish swimming, smoke, cloud and so on. These videos are labeled as 36 classes and each class has 100 subsequences (totally 3600 subsequences) with a fixed size of $50\times 50\times 50$ (50 gray frames). This is a challenging database for clustering because most of texture from different class is fairly similar. Figure \ref{DynTexfig} shows some texture samples.

\begin{figure}
    \begin{center}
    \includegraphics[width=0.45\textwidth,height=25mm]{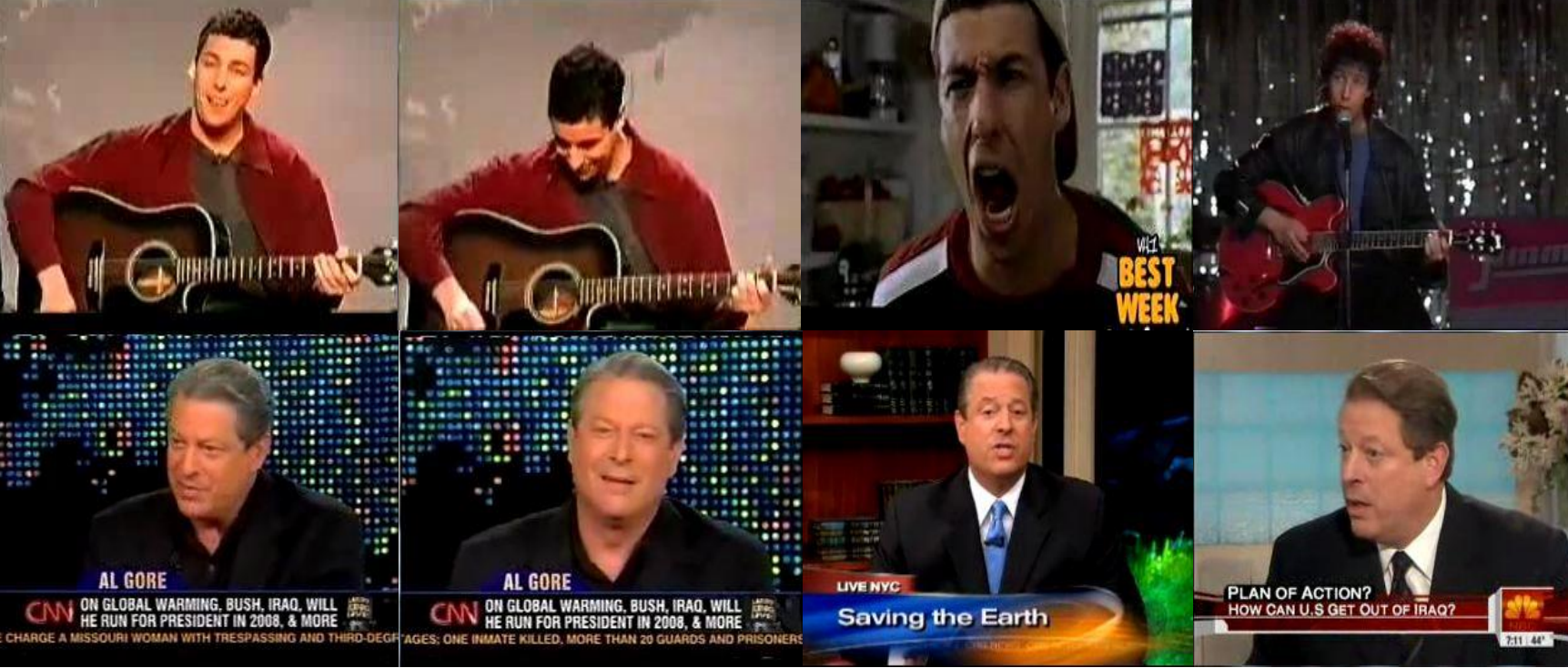}
    \end{center}
    \caption{YouTube Celebrity samples. Each row includes frames from different video sequence of the same person.}\label{YTCfig}
\end{figure}

Finally YouTube Celebrity (YTC) dataset \cite{KimKumarPavlovicRowley2008,WangGuoDavisDai2012} is chosen to test the performance of the proposed method on data with complex variation. YTC, downloaded from Youtube, contains videos of celebrities' joining activities under real-life scenarios in various environments, such as news interviews, concerts, films and so on. The dataset is comprised of 1,910 video clips of 47 subjects and each clip has more than 100 frames. It is a quite challenging dataset since the faces are of low resolution and contains different head pose rotations and rich expression variations. Some samples of YTC dataset are shown in Figure \ref{YTCfig}.

Our strategy is to represent the image set or videos as subspaces objects, i.e., points on Grassmann manifold.
Then a representative basis of such a subspace can be simply obtained by using SVD from the matrix of the raw features of image sets or videos as done in \cite{HarandiSandersonShenLovell2013,HarandiSandersonShiraziLovell2011}. Concretely, given a subset of images, e.g., the same digits written by the same person, denoted by $\{Y_i\}_{i=1}^M$ and each $Y_i$ is a grey-scale image with dimension $m \times n$, we can construct a matrix $\Gamma=[\text{vec}(Y_1), \text{vec}(Y_2),..., \text{vec}(Y_M)]$ of size $(m\times n)\times M$ by vectorizing each image $Y_i$. Then $\Gamma$ is decomposed by SVD as $\Gamma=U\Sigma V$. We pick up the left $p$ singular-vectors of $U$ as the representative of a point on Grassmann  manifold $\mathcal{G}(p,m\times n)$.

In all the experiments, we use both K-means and NCut for the final clustering step over the learned similarity matrix. For the sake of clarity, we summarize all the compared methods in Table \ref{Modeltab}. Under LRR model, the performance of K-means is bad, so we abandon the 4th and last experiment.
$\lambda$ is an important parameter for balancing the error term and the low-rank term of the LRR model on Grassmann Manifold in (\ref{25August2014-4}).
Empirically, we can make $\lambda$ small when the noise of data is lower and use a larger $\lambda$ value if the noise level is higher.

All experiments are implemented by Matlab R2011b and performed on an Xeon-X5675 3.06GHz CPU machine with 12G RAM.
\begin{table*}
  \centering
   \begin{tabular}{|c|c|c|c|c|}
     \hline
              Data processing  & Data representation   & Clustering method & Combining method \\
              method  & method  &  & \\
     \hline
              PCA &  -  & NCut & PNCut \\
     \hline
              PCA &  -  & K-means & PK-means \\
     \hline
              PCA &  LRR  & NCut & PLRRNCut \\
     \hline
              PCA &  LRR  & K-means & - \\
     \hline
              Grassmann Manifold &  - & NCut & GNCut\\
     \hline
              Grassmann Manifold & -  & K-means & GKmeans \\
     \hline
              Grassmann Manifold & LRR  & NCut & GLRRNCut\textbf{(our method)} \\
     \hline
              Grassmann Manifold &  LRR  & K-means & - \\
     \hline

   \end{tabular}
  \caption{Different combining clustering methods with variety in data processing, data representation and clustering methods.}\label{Modeltab}
\end{table*}

\subsection{MNIST Handwritten digits Clustering}
 
To build up subspaces, we create subgroups randomly according to their classes so that the each subgroup consists of 20 images, i.e. $M=20$. We randomly select 40 subgroups for each digit, so our task is to cluster $N=400$ image subgroups into ten categories. Then construct a Grassmann point for each subgroup with $p=10$ as mentioned above. Finally LRR model on Grassmann Manifold is conducted on these points and the result of $Z$ is pipelined to NCut algorithm. As a benchmark comparison, we mimic a point in Euclidean for each subgroup of 20 images by stacking them into a vector of dimension $28\times 28\times 20$. For PCA based methods, the size $28\times28\times20=15680$ of each vector is reduced to $323$ by PCA.

The experimental results are reported in Table \ref{MNISTtab}. It is shown that our proposed algorithm has the highest accuracy of $98\%$, outperforming other methods more than 10 percents. Both GNCut and GK-means methods failed in this experiment. Our explanation is that, after the Grassmann Manifold mapping, it is inappropriate to simply use Euclidean distance as a metric for points on the manifold space. As our LRR model incorporates the geometry over Grassmann manifold, the clustering results are much better than that of PNCut or PLRRNCut, in which Euclidean distance is used to measure the relation of reduced data under PCA. The manifold mapping extracts more useful information about the differences among sample data. Thus the combination of Grassmann geometry and LRR model brings better accuracy for NCut clustering. In our experiments, we set $\lambda=0.3$ for the our LRR model and set $\lambda=0.5$ for PLRR method.

\begin{table*}
  \centering
   \begin{tabular}{|c|c|c|c|c|c|c|}
     \hline
                & PK-means & PNCut & PLRRNCut & GK-means & GNCut & \textbf{Our method}\\
     \hline
     MNIST database     & 0.71 & 0.8675 & 0.8552 & 0.47 & 0.385  & \textbf{0.9833}\\
     YTC  database     & 0.6378 & 0.5189 & 0.2020 & 0.3568 & 0.3568  & \textbf{0.6919}\\
     \hline
   \end{tabular}
  \caption{Subspace clustering results on MNIST and YTC databases.}\label{MNISTtab}
\end{table*}

\subsection{Dynamic Texture Clustering}

As video sequences contain useful space and time context information for clustering, instead of using SVD method, we use Local Binary Patterns from Three Orthogonal Plans (LBP-TOP) model \cite{ZhaoPietikaeinen2007} to construct points on Grassmann Manifold. The LBP-TOP method extracts LBP features from three orthogonal planes and concatenate a number of neighbour points' features to form a co-occurrence feature. After extracting LBP-TOP features for the 3600 subsequences, we get 3600 matrices of size $177\times 14$ as the points on Grassmann Manifold.
As the data volume of all the 3600 subsequences is huge, we randomly pick $K (=3,...,10)$ classes from 36 classes and 50 subsequences for each class to cluster. The experiments are repeated several times for each $K$. In the experiments, $\lambda$ is also set to 0.8. For the PCA based method, the prototype $50\times 50\times 50$ subsequence is reduced to 221 and $\lambda$ for the PLRRNCut is set to 0.6.

\begin{figure}
    \begin{center}
    \includegraphics[width=0.45\textwidth]{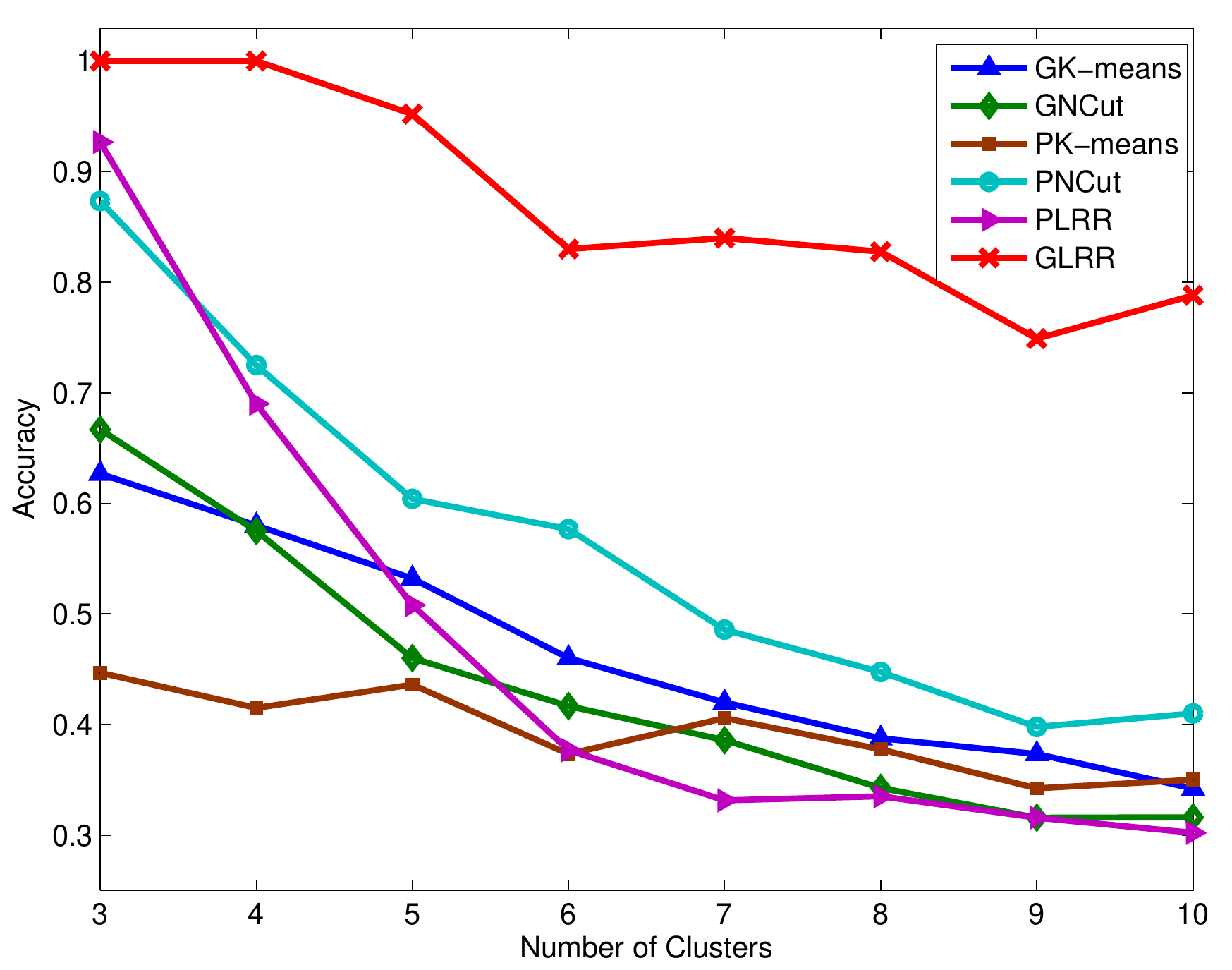}
    \end{center}
    \caption{Clustering results on DynTex++ database with the number of classes ranging from 3 to 10.}\label{DyntexAcc}
\end{figure}

The clustering results for DynTex++ database are shown in Figure \ref{DyntexAcc}. For different number of classes, the accuracy of the proposed method is superior to the other methods more than 10 percents, due to incorporating manifold information over the LBP-TOP features. We also observe that all of accuracies decreases as the number of classes increases. This may be caused by the clustering challenge when more similar texture images are added into the data set.

\subsection{YouTube Celebrity Clustering}
 
In order to create a face dataset form the YTC videos, a face detection algorithm is exploited to extract face regions and resize each face to a $20\times 20$ image. We treat the faces extracted from each video as an image set without thinking about the number $M$ of frames in different videos. LBP-TOP is not appropriate for the extracted face images in this case as we have lost relative spatial information,  
so we use SVD method  to construct points on Grassmann Manifold like the handwritten database. We only test the algorithm on $K=4$ classes with $N=218$ video clips as an example for this complicated data case. For the PCA based method, the size of each subsequence is reduced to 3000 and $\lambda$ for the PLRRNCut is set to 0.6.

The clustering results for YTC face dataset is shown in Table \ref{MNISTtab}. The accuracy of our method is significantly higher than other methods although it is relatively lower than the accuracy for DynTex dataset. One of key reasons for this lower accuracy may be due to rapid changes existed in face images from a clip.

\section{Conclusion}\label{Sec:6}
In this paper, building on Grassmann manifold geometry and the Log mapping, the classic LRR model is extended for object subspaces in Grassmann manifold. The data self expression used in LRR is implemented over the tangent spaces at points on Grassmann manifold, thus a novel LRR on Grassmann manifold, namely GLRR, is presented. An efficient algorithm is also proposed for the GLRR model. Our experiments on image sets and video clip databases confirm that GLLR is well suitable for representing non-linear high dimensional data and revealing intrinsic multiple subspaces structures in clustering applications. The experiments also demonstrate that the proposed method behaves effectively and steadily in variety of complicated practical scenarios. As part of future work, we will explore the LRR model on Grassmann manifold in an intrinsic view.

\section*{Acknowledgements}
The research project is supported by the Australian Research Council (ARC) through the grant DP130100364 and also partially supported by National Natural Science Foundation of China under Grant No. 61390510, 61133003, 61370119, 61171169, 61300065 and Beijing Natural Science Foundation No. 4132013.

{\small
\bibliographystyle{ieee}
\bibliography{reference_boyue}
}

\end{document}